\documentclass{article}

\usepackage{microtype}
\usepackage{graphicx}
\usepackage{subfigure}
\usepackage{booktabs} 
\usepackage{multirow}
\usepackage{array}
\usepackage{amsmath}
\usepackage{amssymb}
\usepackage{mathtools}
\usepackage{amsthm}
\usepackage{bbm}
\usepackage{natbib}
\usepackage{tikz}
\newcommand*\circled[1]{\tikz[baseline=(char.base)]{
            \node[shape=circle,draw,inner sep=2pt] (char) {#1};}}

\usepackage[accepted]{icml2024arxiv}

\usepackage{hyperref}

\newcommand{\bE}{\mathbb{E}}

\newcommand{\cG}{\mathcal{G}}
\newcommand{\cD}{\mathcal{D}}
\newcommand{\cX}{\mathcal{X}}
\newcommand{\MSE}{\textnormal{MSE}}
\newcommand{\ASCE}{\textnormal{ASCE}}
\newcommand{\ECE}{\textnormal{ECE}}
\newcommand{\gASCE}{\textnormal{gASCE}}

\newcommand{\Accuracy}{\textnormal{Accuracy}}
\newcommand{\Confidence}{\textnormal{Confidence}}

\newcommand{\Var}{\textnormal{Var}}
\newcommand{\Py}{\mathbb{P}}
\newcommand{\hf}{\hat{f}}

\newcommand{\grid}{[\tfrac{1}{m}]}
\usepackage[textsize=tiny]{todonotes}
\DeclareMathOperator*{\argmax}{arg\,max}
\DeclareMathOperator*{\argmin}{arg\,min}

\definecolor{editone}{HTML}{0000FF}

\usepackage{color-edits}
\addauthor{mb}{teal}

\usepackage[capitalize,noabbrev]{cleveref}

\theoremstyle{plain}
\newtheorem{theorem}{Theorem}[section]
\newtheorem{proposition}[theorem]{Proposition}

\theoremstyle{definition}
\newtheorem{definition}[theorem]{Definition}

\theoremstyle{remark}

\icmltitlerunning{Multicalibration for Confidence Scoring in LLMs}

\begin{document}

\twocolumn[
\icmltitle{Multicalibration for Confidence Scoring in LLMs}



\icmlsetsymbol{equal}{*}

\begin{icmlauthorlist}
\icmlauthor{Gianluca Detommaso}{a}
\icmlauthor{Martin Bertran}{equal,a}
\icmlauthor{Riccardo Fogliato}{equal,a}
\icmlauthor{Aaron Roth}{a,b}
\end{icmlauthorlist}

\icmlaffiliation{a}{AWS AI}
\icmlaffiliation{b}{University of Pennsylvania}

\icmlcorrespondingauthor{Gianluca Detommaso}{detommaso.gianluca@gmail.com}

\icmlkeywords{multicalibration, LLMs}

\vskip 0.3in
]



\printAffiliationsAndNotice{\icmlEqualContribution} 

\begin{abstract}
This paper proposes the use of ``multicalibration'' to yield interpretable and reliable confidence scores for outputs generated by large language models (LLMs). Multicalibration asks for calibration not just marginally, but simultaneously across various intersecting groupings of the data. We show how to form groupings for prompt/completion pairs that are correlated with the probability of correctness via two techniques: clustering within an embedding space, and ``self-annotation'' --- querying the LLM by asking it various yes-or-no questions about the prompt. We also develop novel variants of multicalibration algorithms that offer performance improvements by reducing their tendency to overfit.  Through systematic benchmarking across various question answering datasets and LLMs, we show how our techniques can yield confidence scores that provide substantial improvements in fine-grained measures of both calibration and accuracy compared to existing methods. 
\end{abstract}

\section{Introduction}
Large language models (LLMs) have revolutionized text generation, with applications ranging from code development \cite{chen2021evaluating} to information retrieval \cite{zhu2023large}. However, alongside their impressive capabilities, LLMs possess a troubling tendency to fabricate information, generating outputs that diverge from factual reality – a phenomenon dubbed ``hallucination'' \cite{huang2023survey}. These hallucinations pose significant challenges to the trustworthiness and ethical deployment of LLMs, demanding the development of robust detection and mitigation strategies.
\begin{figure}[t!]
    \centering
    \includegraphics[width=\columnwidth]{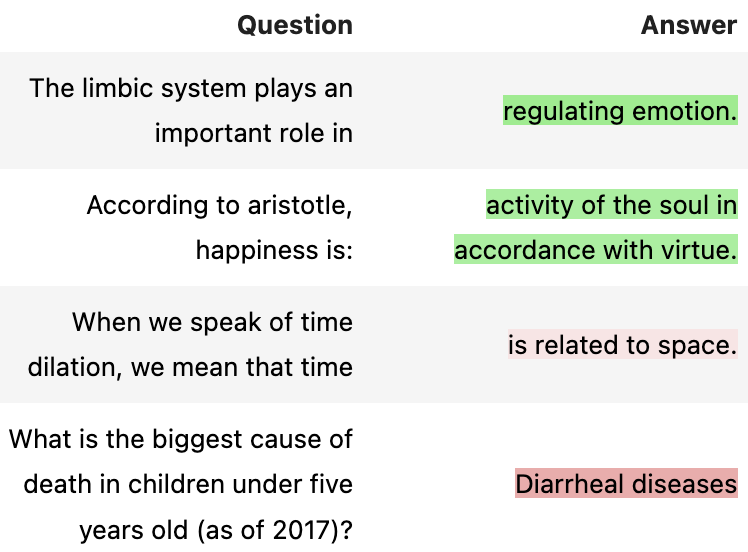}
    \caption{An application of multicalibration to question answering. Answers are colored from red to green according to their multicalibrated confidence scores of being a hallucination. Multicalibration is performed using Algorithm \ref{alg:iglb}.}
    \label{fig:color}
    \vspace{-0.3cm}
\end{figure}

In this paper, we leverage recent ``multicalibration'' techniques \cite{hebert2018multicalibration} to produce calibrated probabilities indicating whether a generated response constitutes a hallucination. Unlike conventional calibration methods, multicalibrated probabilities are self-consistent not just marginally (i.e. on average over all examples), but also \emph{conditionally} on various properties of the instance, which allows them to serve as more refined risk measures. Producing ``risk scores'' for hallucinations can provide an interpretable measure of risk which can be exposed to the user 
(e.g. through a coloring scheme, as in Figure \ref{fig:color})
to communicate the risk associated with the generated content. Moreover, when those risk scores are calibrated, they are not only interpretable but ``trustworthy'' in the sense that they can be safely used as if they were true probabilities \cite{noarov2023high}. 

Our approach mirrors the robust assurances offered by conformal prediction, where multicalibration (of quantiles) has been used to give group-conditional guarantees \cite{bastani2022practical,jung2022batch,gibbs2023conformal}. Traditionally multicalibration has been used to give estimates of uncertainty in tabular data settings that hold conditionally on various features that are explicitly present in the data --- often demographic attributes like sex or race. A key challenge in applying these techniques to hallucination detection in LLMs is a lack of such explicit features. An important part of our contribution is generating features that are useful to multicalibrate with respect to --- which we do both through  clustering prompt embeddings, and by having the LLM itself annotate prompts with binary features via the answers to yes-or-no questions. 

We note that what in many contexts, what is  and is not a ''hallucination`` can be open to interpretation, and does not have sharp boundaries. In this study, we adopt an agnostic stance toward its definition. Specifically, we refrain from stipulating criteria for determining what constitutes "good" or "bad" generated content. Instead, we assume access to a modestly sized calibration dataset that has been annotated with binary labels. For any criterion for what constitutes a ``good'' vs. ``bad'' completion in a given context, such a dataset could be produced by human evaluators. For our work, we assume that this is given and do not take a stance on what the criterion for establishing that a given completion is ``good'' in a given context should be.

Our contributions are threefold: 1.~We show how to apply multicalibration techniques in the context of hallucination detection in LLMs; a primary challenge here is to obtain reasonable ``groups'' with respect which to multicalibrate, which we do via prompt clustering and via self-annotation of prompts. 2.~We introduce novel variations of multicalibration methods which yield substantial performance enhancements. 3.~We systematically evaluate these techniques across diverse LLMs and question answering datasets, demonstrating their efficacy in calibration and overall performance compared to existing baselines.

\vspace{-.15in}
\paragraph{Additional Related Work}
Numerous recent surveys focus on hallucinations in LLMs \cite{chang2023survey, huang2023survey, ji2023survey, rawte2023survey, tonmoy2024comprehensive, zhang2023siren, guerreiro2023hallucinations}. The predominant focus of current research lies in binary hallucination \textit{detection}, specifically the capacity to discern whether generated text exhibits signs of hallucination. Key contributions in this domain include \cite{manakul2023selfcheckgpt, rebedea2023nemo}, which evaluate consistency, similarity, and agreement among alternative generated responses. \citep{kadavath2022language, friel2023chainpoll} directly engage LLMs by posing inquiries about correctness or consistency within a single answer.

More closely related is a smaller body of literature that  explores uncertainty quantification and confidence scoring in this context \cite{xiao2021hallucination, verma2023reducing, varshney2023stitch, kalai2023calibrated, tian2023just, zhao2023automatic, chen2023quantifying, duan2023shifting, lin2023generating, liu2023calibrating} and propose a variety of approaches to reduce hallucination generation ranging from updated beam search methods, fine-tuning, human labelling, and epistemic neural networks. Several recent papers use conformal prediction to derive sets of prompt completions, offering marginal coverage guarantees (e.g.~for 90\% of prompts, at least one completion in the set should be ``good'') \cite{quach2023conformal, kumar2023conformal, deutschmann2023conformal, ren2023robots, zecchin2023forking}. Among these, \cite{kumar2023conformal} is closest to our approach but requires "group-specific" prompting strategies. The remainder focus on improving the LLM's decoding strategy and/or predictive sets, which are less suited to binary classification settings (like hallucination detection), where they are limited to $\{0\}$, $\{1\}$, and $\{0, 1\}$.


\section{Background on (Multi)Calibration}\label{sec:calibration}

Consider $(X, Y)\sim \cD$ where $X\in \cX$ indicates a prompt/completion pair, $Y$ indicates whether the completion is a hallucination given the prompt ($Y=0$) or not $(Y=1)$, and $\cD$ represents the joint distribution over pairs $(X,Y)$. Let $f:\cX\mapsto[0,1]$ denote a  score representing a confidence that $Y=1$ for the text $X$. See Section \ref{sec:scoring} for a discussion about possible scores $f(x)$. 

Ideally, we would like to find a model $f(x)$ such that
\begin{equation}\label{eq:ideal_f}
    f(x) = \Py_\cD(Y=1|X=x),\quad\forall\ x\in\cX.
\end{equation}
However, there are two difficulties with this. First, this may not be a coherent probabilistic notion: fixing $x$, the label may be determined, and so the ``probability'' of $Y=1$ may be either $0$ or $1$. Moreover, it is generally impossible to learn a function with this property without observing \emph{every} possible $x$, which is impossible for extremely large sets $\cX$, as is the case for LLM prompt/completion pairs \cite{lei2014distribution}. 
Calibration is a simple, tractable, guarantee that corresponds to a significant coarsening of the set of conditioning events in Equation~\ref{eq:ideal_f}, to the level sets of the predictor $f$.

\begin{definition}[Calibration]\label{def:calib}
Given a data distribution $\cD$, the \textit{bias} of a model $f$ at the $p$-th level set is defined as
\begin{equation}\label{eq:bias}
    \Delta_p(f) := \bE_\cD[Y - f(X)|f(X)=p].
\end{equation}
Then, if $\Delta_p(f)=0$ for all $p\in [0, 1]$ such that $\Py_\cD(f(X)=p)>0$, we say that $f$ is \textit{calibrated} w.r.t.~$\cD$.
\end{definition}
We observe that Definition \ref{def:calib} can be rewritten as 
\[ \Py_\cD(Y=1|f(X)=p)=p. \]
Informally, calibration is a minimal consistency condition: it states that the conditional distribution on $Y$ conditional on the prediction that $f(X) = p$ is indeed a Bernoulli distribution with bias $p$.  While a perfect model satisfying $\eqref{eq:ideal_f}$ is calibrated, the converse is not necessarily true.

We introduce the following standard measure of calibration error. 
\begin{definition}[Average squared calibration error]\label{def:asce}
    We denote the \textit{average squared calibration error} (ASCE) of a model $f$ w.r.t.~a distribution $\cD$ by
    \begin{equation}\label{eq:asce}
        \ASCE(f) := \bE_P[\Delta_P^2(f)].
    \end{equation}
\end{definition}
The ASCE is computed by integrating the squared model bias over all level sets. Note that if the model $f$ is calibrated, then $\ASCE(f)=0$. The ASCE is related to the well-known expected calibration error (ECE) \cite{naeini2015obtaining}, with the difference that the ASCE compares $Y$ against $f$, while the ECE compares accuracy against confidence (see Appendix \ref{app:ECE}). ASCE is a useful measure of calibration error because it is directly related to a natural measure of accuracy: mean squared error.
\begin{definition}[Mean squared error]
    We denote the \textit{mean squared error} (MSE) of a model $f$ w.r.t.~a distribution $\cD$ by
    \begin{equation}
        \MSE(f) := \bE_{\cD}\left[(Y - f(X))^2\right].
    \end{equation}
\end{definition}
The MSE is also known as \textit{Brier score} \cite{brier1950verification}. A bias-variance decomposition clarifies its relation to the ASCE.
\begin{proposition}[See, e.g.~\cite{kohavi1996bias} ]\label{prop:mse_decomp}
We have
\begin{equation}
    \MSE(f) = \ASCE(f) + \bE_P[\Var_{\cD}(Y|f(X)=P)]. 
\end{equation}
\end{proposition}
A proof is given in Appendix \ref{app:proof_mse_decomp} for completeness. Proposition \ref{prop:mse_decomp} shows that the MSE can be decomposed as the ASCE and the  variance of the data given the model, respectively measuring how calibrated the model is and how much variation in the data the model can explain. As such, the MSE is not a direct measure of calibration --- when comparing two models, the model with lower squared error may still be the less well calibrated model. 

\subsection{A Simple Calibration Strategy}\label{sec:model_patch}
\begin{algorithm}[h!]
  \caption{Histogram Binning (HB)}
  \label{alg:hb}
  \begin{algorithmic}[1]
    \FORALL{$p\in\grid$}
    \STATE Set 
    \vspace{-0.5cm}
\begin{equation}
    \hf(x):= \begin{cases}
    f'(x) + \Delta_p(f') &\mbox{if } x\in S_p(f'),\\
    f'(x) &\mbox{otherwise.}
    \end{cases}
\end{equation}
\vspace{-0.5cm}
    \ENDFOR
    \end{algorithmic}
\end{algorithm}
Given a distribution $\cD$, a model $f$, and a threshold $\alpha>0$, our goal is to produce a new model $\hf$ that is calibrated and has reduced MSE compared to $f$ --- at least up to discretization error $\alpha$. Here $\alpha$ will control the number of level sets we discretize $f$ to, and hence the complexity of the model: choosing smaller values of $\alpha$ will require more data to avoid overfitting and vice versa. First, we introduce the notation for the level sets which appear as conditioning events in \eqref{eq:bias}:  $S_p(f):=\{f(x)=p\}$. Since it is infeasible to condition on $S_p(f)$ for all $p\in[0, 1]$, we introduce the uniform grid $\grid:=\{\tfrac{i}{m}\}_{i=0}^m$, and define
\begin{equation}
    f'(x) := \argmin_{p\in\grid}|f(x) - p|,
\end{equation}
which rounds the model $f$ to the grid $\grid$. Algorithm \ref{alg:hb} rounds the model, then applies a constant shift in the bin $S_p(f')$, for each element $p$ in the grid. Because $\Delta_p(\hf)=0$ for all $p\in\grid$, the following result holds.
\begin{theorem}[See, e.g.~\cite{roth2022uncertain}]\label{thm:level set}
Algorithm \ref{alg:hb} satisfies $\ASCE(\hf)=0$. Furthermore, set $B_p:=\{x : |f(x)-p|\le\tfrac{1}{2m}\}$. If $m>\tfrac{1}{\alpha}$, then
\vspace{-0.2cm}
\[ \MSE(\hf) < \MSE(f) - \sum_{p\in\grid}\Py_{\cD}\left(B_p\right)\Delta_p^2(f') + \frac{\alpha^2}{4} + \alpha.\]
\end{theorem}
\vspace{-0.2cm}
A proof is given in Appendix \ref{app:proof_thm_level set} for completeness. Theorem \ref{thm:level set} shows that by replacing the level sets of the model (on a refined enough grid) with the label mean of points conditional on the level set not only guarantees calibration, but improves the MSE of the model by its initial calibration error (as measured by ASCE). This is a key property of calibration and related guarantees: enforcing it is only accuracy enhancing. Furthermore, one can prove out-of-sample generalization bounds that replace the joint distribution $\cD$ with the empirical distribution characterized by the available data, and hence satisfy a non-asymptotic calibration guarantee --- see e.g.  \cite{roth2022uncertain}. For simplicity, and because the generalization bounds are standard, in our exposition here we will focus on in-sample guarantees. As Algorithm \ref{alg:hb} is closely related to histogram binning \cite{zadrozny2001obtaining}, we will refer to it with this name.

HB crucially uses the fact that  the level sets $S_p(f)$ defined in Definition \ref{def:calib} are disjoint. In the following section, we will define multicalibration \cite{hebert2018multicalibration}, a stronger calibration guarantee that imposes simultaneous requirements on non-disjoint conditioning sets.

\subsection{Towards Multicalibration}\label{sec:towards_multivalid}
In this section, we argue that the promise made by calibration in Definition \ref{def:calib} is too weak for the kinds of language model applications we have in mind because the performance of a model is extremely heterogeneous across different kinds of tasks that it can be used for. As an example, consider two prompt/completion pairs $x_1$ and $x_2$, respectively asking and answering a question about (1) the capital cities of US states, and (2)  citations to the academic literature for theorems in functional analysis. We would expect that the  probability of correctness differs substantially across these examples --- and yet calibration is a \emph{marginal} guarantee that can average over both cases. It could, for example, lead to confidence assessments that are systematically over-confident about academic citations and systematically under-confident about state capitals. This is not in conflict with even perfect calibration. It would be better to promise that our confidence scores were calibrated conditionally on (as fine-grained information as possible about) the prompt used. These kinds of conditional calibration guarantees are what multicalibration aims for.

We now formalize the concept of \textit{groups}. A group function $g:\mathcal{X}\to\{0, 1\}$ can be thought of as an indicator function for a group defined as a set of prompt/completion pairs: $g(x) = 1$ if $x$ is a member of the ``group'' and $g(x) = 0$ otherwise. The ``group'' induced by an indicator function $g$ is therefore $\{x \in \cX : g(x)=1\}$. A set of groups $\cG$ is correspondingly identified by a set of group indicator functions. Crucially, the groups in a collection $\cG$ can be \emph{intersecting} --- i.e. there can be multiple groups that contain the same example $x$. This corresponds to a prompt/completion pair having multiple non-mutually-exclusive attributes: for example, $x$ might simultaneously be ``a question requiring high-school level knowledge'' and ``a question about mathematics''.

\subsubsection{Group-Conditional Unbiasedness}\label{sec:gcu}
\begin{algorithm}[h!]
  \caption{Group-Conditional Unbiased Regression}
  \label{alg:gcur}
  \begin{algorithmic}[1]
    \STATE Set 
    \vspace{-0.5cm}
    \begin{equation}\label{eq:gcu_model}
\begin{split}
    &\hat{f}(x) := f(x) + \sum_{g\in\cG}\lambda_g\, g(x),\\
    &\text{s.t.}\ \{\lambda_g\}_{g\in\cG} = \argmin \MSE(\hf).
\end{split}
\end{equation}
\vspace{-0.5cm}
    \end{algorithmic}
\end{algorithm}

Before introducing multicalibration, let us first introduce a simpler guarantee, and a simple strategy to obtain it. Calibration (Definition \ref{def:calib}) requires that a model $f$ be unbiased conditional on its own level sets. Given a collection of groups $\cG$, we can instead ask a model to be unbiased conditionally on each of the group indicator functions in $\cG$.  

\begin{definition}[Group-conditional unbiasedness]
Given a data distribution $\cD$ and set of groups $\cG$, we say that a model $f$ is \textit{group-conditionally unbiased} if
\begin{equation}
    \bE_\cD[Y - f(X)|g(X)=1]=0,\quad\forall\,g\in\cG.
\end{equation}
\end{definition}
This condition is also known as ``multi-accuracy'' in the algorithmic fairness literature \cite{hebert2018multicalibration,kim2019multiaccuracy}.

Since groups may be overlapping, we cannot proceed as in Section \ref{sec:model_patch} and independently unbias the predictions within each group. Instead, we introduce Algorithm \ref{alg:gcur}, initially proposed by \cite{gopalan2022low}, in which a model $\hf$ is fit by solving a linear regression problem over features defined both by the original model $f$ and the group indicator functions in $\cG$. The following theorem is due to \cite{gopalan2022low}. We follow the presentation of \cite{roth2022uncertain}. 
\begin{theorem}\label{thm:gcu}
    The model $\hf$ produced in   \eqref{eq:gcu_model} satisfies group-conditional unbiasedness.
\end{theorem}
A proof is given in Appendix \ref{app:thm_gcu} for completeness. Once again, with standard techniques one can prove generalization bounds for this algorithm (see \cite{roth2022uncertain}) which allows one to replace in-sample MSE with true distributional MSE; we elide these details for simplicity here.  The model produced in \eqref{eq:gcu_model} is solving a linear regression problem (minimizing squared error over a set of linear models). In Appendix \ref{app:thm_gcu}, we generalize this result to show that it holds also when we replace MSE with a cross-entropy loss --- i.e. when solving logistic regression rather than linear regression. We name the latter method \textit{Group-Conditional Unbiased Logistic Regression} (GCULR). A similar generalization is given by \cite{gopalan2023loss}.

\subsubsection{Multicalibration}\label{sec:multivalid_calib}
In Section \ref{sec:calibration}, we defined calibration by conditioning on the level sets of the model. In Section \ref{sec:gcu}, we defined group-conditional unbiasedness by conditioning on a set of groups. Multicalibration, introduced by \cite{hebert2018multicalibration}, is a stronger guarantee that asks for unbiasedness when simultaneously conditioning on both level sets and groups. In order to rigorously define it, let us first generalize the definition of $\ASCE$.
\begin{definition}[Group average squared calibration error]\label{def:gasce}
    Given a group function $g$, we denote the \textit{group average squared calibration error} (gASCE) of a model $f$ w.r.t.~a distribution $\cD$ by
    \begin{equation}\label{eq:gasce}
    \gASCE(f, g) := \bE_P\hspace{-2pt}\left[\Delta_{P, g}^2(f)|g(X)=1\right],
    \end{equation}
    where $\Delta_{p, g}(f):=\bE_\cD[Y - f(X)|S_{p,g}(f)]$ and $S_{p,g}(f):=\{f(x)=p, g(x)=1\}$.
\end{definition}

Unlike the $\ASCE$, in the $\gASCE$ we do not only condition on the disjoint level sets $\{f(x)=p\}$, but also on the groups $\{g(x)=1\}$, which allows us to quantify the calibration error independently for each group.
\begin{definition}[Multicalibration]\label{def:multivalid}
Given a data distribution $\cD$ and a set of groups $\cG$, a model $f$ is $\alpha$-approximately \textit{multicalibrated} w.r.t.~$\cD$ and $\cG$ if and only if
\begin{equation}\label{eq:gasce}
    \gASCE(f, g)<\tfrac{\alpha}{\Py_{\cD}(g(X)=1)},\quad \forall\,g\in\cG.
\end{equation}
\end{definition}
This is an $\ell_2$ notion of multicalibration, as studied in \cite{Globus-HarrisHK23}. One can also study error in other metrics (e.g. $\ell_\infty$ error as in \cite{hebert2018multicalibration} or $\ell_1$ error as in \cite{gopalan2022omnipredictors}). 

Compared to the Definition \ref{def:calib} of calibration, multicalibration is a stronger guarantee, where standard calibration is recovered by taking $\cG=\{\cX\}$. 
Because the groups may overlap, we cannot independently apply patches for all conditioning sets. However, one can use a similar idea with an iterative approach, which results in an algorithm that is guaranteed to satisfy multicalibration in a finite number of rounds, and decrease the MSE at every round.

\begin{algorithm}
  \caption{Iterative Grouped Histogram Binning (IGHB)}
  \label{alg:ighb}
  \begin{algorithmic}[1]
    \STATE Let $m=\lceil\tfrac{1}{\alpha}\rceil$, $t=0$, $f_0:=f'$.
    \WHILE{$\max_{g\in\cG}\,\Py_{\cD}(g(X)=1)\,\gASCE(f, g)>\alpha$}
    \STATE Set 
    \[(p_t, g_t) = \argmax_{p\in\grid,\, g\in\cG}\Py_{\cD}\left(S_{p, g}(f_t)\right)\,\Delta_{p, g}^2(f_t).\]
    \STATE Set 
    \begingroup\makeatletter\def\f@size{8.5}\check@mathfonts
\def\maketag@@@#1{\hbod{\m@th\normalsize\normalfont#1}}$
    h_{t+1}(x):=\begin{cases}f_t(x) + \Delta_{p_t, g_t}(f_t) &\mbox{if $x\in S_{p_t,g_t}(f_t)$},\\
    f_t(x) &\mbox{otherwise.}\end{cases}
    $.
    \endgroup
    \STATE Set $f_{t+1}:=h_{t+1}'$.
    \ENDWHILE
    \end{algorithmic}
\end{algorithm}

IGHB (Algorithm \ref{alg:ighb}) (a variant of which was first given by \cite{hebert2018multicalibration}) starts by checking whether $\alpha$-approximate multicalibration is satisfied. If not, it finds the conditioning event for which the $\gASCE$ is largest, and patches the model on examples that trigger that event. It iterates like this until convergence. The rounding operation makes sure that the number of level sets do not increase without bound, which guarantees that there is sufficient data to evaluate the bias on each of the conditioning events.

\begin{theorem}\label{thm:multivalid}
    Algorithm \ref{alg:ighb} halts after $T < \tfrac{4}{\alpha^2}$ rounds and returns a model $f_T$ that is $\alpha$-approximately multicalibrated. Moreover, if the algorithm runs for $T$ rounds, then 
    \[ \MSE(f_T) < \MSE(f) - (T-1)\tfrac{\alpha^2}{4} + \alpha.\vspace{-0.2cm}\]
\end{theorem}
A proof can be found in \cite{roth2022uncertain} along with out-of-sample generalization guarantees. As with HB, running IGHB is only accuracy-improving.

\section{Remedying overfitting}\label{sec:overfitting}
The multicalibration strategies outlined in \ref{sec:multivalid_calib} can build complex models (because of iterative updates on intersecting groups), and are known to be subject to overfitting in practice (see e.g. \cite{Globus-HarrisHK23}). One reason for this is that the technique operates by iteratively estimating the label mean on subsets of the data defined as  $\{f(x)=v, g(x)=1\}$, which in-sample can contain very few points and thus lead to inaccurate estimates of distributional quantities. Here, we provide practical improvements to the IGHB algorithm that lead to better performance.

\subsection{Bins as upper and lower sets}\label{sec:uplowsets}
We make the following observation: rather than conditioning on the exact value of the model $f(x) = p$, we can condition on $f(x) \leq p$ (roughly speaking conditioning on values of its CDF rather than its density function), and the definition of (exact) multicalibration remains unchanged.

\begin{proposition}\label{prop:equiv_multivalid}
If $\bE_\cD[Y|f(X)=p]$ is a continuous function of $p$, the definition of perfect multicalibration (\textnormal{Definition} \ref{def:multivalid} with $\alpha = 0$) is unchanged if we replace $S_{p, g}$ by $S^{\le}_{p, g}:=\{f(x)\le p, g(x)=1\}$.
\end{proposition}
A proof is in Appendix \ref{app:equiv_multivalid}. Proposition \ref{prop:equiv_multivalid} suggests a definition of multicalibration that is defined on considerably larger conditioning sets. By symmetry, the same results holds for $S^{\ge}_{p, g}:=\{f(x)\ge p, g(x)=1\}$. Thus ``patching'' sets $S^{\ge}_{p, g}$ or $S^{\le}_{p, g}$ within the IGHB algorithm when the model exhibits bias on them moves the model closer to multicalibration, and the same convergence analysis applies. Note that $S_{p, g}^{\le}$ is larger for large $p$, and vice versa for $S_{p, g}^{\ge}$, hence one may want to use one or the other bin according to the value of $p$. Note that the per-bin calibration error in step 3 of Algorithm \ref{alg:ighb} is proportional to the size of the bin, implying that larger bins are likely to be patched earlier than smaller bins. Updates on large sets are less prone to overfitting because we have many samples to use to estimate their label mean. It follows that without invalidating Theorem \ref{thm:multivalid}, in order to patch the model on a sequence of considerably larger bins we can simply replace step 3 in Algorithm \ref{alg:ighb} with
\begin{equation}\label{eq:new_bins}
    (p_t, g_t, \tau_t) =\hspace{-0.3cm} \argmax_{p\in\grid,\, g\in\cG, \tau\in\{\le,\ge\}} \hspace{-0.3cm} \Py_{\cD}\left(S_{p, g}^\tau(f_t)\right)\,(\Delta_{p, g}^\tau(f_t))^2,
\end{equation}
where $\Delta_{p, g}^\tau$ is defined as $\Delta_{p, g}$ in Definition \ref{def:asce} but replacing $S_{p, g}$ with $S_{p, g}^\tau$. To reiterate, what makes this approach work is that (1), a model with no  bias over the sets $S_{p, g}^\tau$ also has no bias over the sets $S_{p, g}$ and hence satisfies multicalibration, and (2) the sets $S_{p, g}^\tau$ are larger and hence reduce overfitting both because updating on larger sets moves the model more quickly to convergence, and estimating distributional parameters on larger sets leads to less error.

\subsection{Linear Scaling}\label{sec:lin_scal}
\begin{algorithm}[h!]
  \caption{Linear Scaling (LS)}
  \label{alg:ls}
  \begin{algorithmic}[1]
    \STATE Set 
    \vspace{-0.5cm}
    \begin{equation}\label{eq:ls}
\begin{split}
    &\text{LS}[f](x) := \text{expit}(\alpha^* + \beta^*\, \text{logit}\,f(x)),\\
    &\text{with}\ (\alpha^*,\beta^*) = \argmin_{\alpha,\beta} \MSE(\hf).
\end{split}
\end{equation}
\vspace{-0.5cm}
    \end{algorithmic}
\end{algorithm}
Since Algorithm \ref{alg:ighb} operates over conditioning events $S_{p, g}$ over which the current model takes constant value, it is reasonable to apply constant patches in the bins that are independent of the model value. However, when we start using conditioning events $S_{p, g}^\tau$ like in \eqref{eq:new_bins}, the model is no longer constant subject to the conditioning events, and it is reasonable to explore alternative updates that depend on the model value. It is immediate to extend the patches in step 4 of Algorithm \ref{alg:ighb} to linear patches of the form $h_{t+1}(x)=\alpha + \beta f_t(x)$ for $x\in S_{p_t, g_t}$, without affecting the results of Theorem \ref{thm:multivalid}, so long as we choose $\alpha$ and $\beta$ so as to minimize the squared error of the model. In fact, if $\alpha=\Delta_{p_t,g_t}(f_t)$ and $\beta=1$, we recover the patch in step 4. This works because the analysis of Algorithm \ref{alg:ighb} involves showing that the MSE of the model decreases at every step; by minimizing MSE over a model class that can represent the patches used in Step 4 of the algorithm, the analyzed convegence only becomes more rapid. In practice, to avoid clipping $f_t$ between 0 and 1, we make use of the logit and expit (a.k.a.~sigmoid) functions, which respectively map $f$ to an unconstrained domain, and then map a linear transformation of it between $0$ and $1$. Concretely, we replace step 4 in Algorithm \ref{alg:ighb} with
\begin{equation}\label{eq:ls_patch}
\vspace{-0.1cm}
    h_{t+1}(x):=\begin{cases}
        \text{LS}[f_t](x) &\mbox{if }x\in S_{p_t, g_t}^{\tau_t}(f_t)\\
        f_t(x) &\mbox{otherwise,}
    \end{cases}
\vspace{-0.1cm}
\end{equation}
where $\text{LS}[f]$ is defined in \eqref{eq:ls}. When applied as a stand-alone calibration method over the whole input space $\cX$, Algorithm \ref{alg:ls} is related to similar methods such as Platt scaling \cite{platt1999probabilistic} and temperature scaling \cite{guo2017calibration}. Instead, we apply it to the conditioning events selected in Algorithm \ref{alg:ighb} with the goal of multicalibration.

\subsection{Early stopping}
 Algorithm \ref{alg:ighb} is an iterative algorithm that builds a model whose complexity grows with the number of iterations it runs for. Hence a natural heuristic for mitigating overfitting is to implement an early stopping rule. To do so, we initially partition the available data into calibration and validation sets. We then halt the algorithm when the MSE on the validation set ceases to decrease. The rationale of this is supported by Theorem \ref{thm:multivalid}, which establishes that the MSE must decrease on the calibration data in every round. A lack of MSE reduction signifies potential overfitting, rendering the MSE a meaningful loss function for early stopping.

Another intuitive criterion for early stopping is to assess whether the probability mass of the conditioning event selected by the algorithm for an update exceeds a specified threshold. Recognizing that patches on small conditioning events may potentially compromise the algorithm's generalization ability, we choose to halt the process whenever a conditioning event of insufficient size is chosen for patching.

Algorithm \ref{alg:iglb} combines the strategies discussed in Section \ref{sec:overfitting}.

\vspace{-0.25cm}
\begin{algorithm}
  \caption{Iterative Grouped Linear Binning (IGLB)}
  \label{alg:iglb}
  \begin{algorithmic}[1]
    \STATE Let $t=0$, $f_0:=f'$, $\varepsilon > 0$. Split $\cD$ into $\cD_{\text{calib}}$ and $\cD_{\text{val}}$.
    \WHILE{True}
    \STATE Set $(p_t, g_t, \tau_t)$ as in \eqref{eq:new_bins}.
    \STATE \textbf{Break} if $\Py_{\cD_{\text{calib}}}\left(S_{p_t, g_t}^{\tau_t}(f_t)\right) < \varepsilon.$
    \STATE Set $h_{t+1}(x)$ as in $\eqref{eq:ls_patch}$.
    \STATE \textbf{Break} if $\MSE_{\cD_{\text{val}}}(h_{t+1}) \ge \MSE_{\cD_{\text{val}}}(f_t)$.
    \STATE Set $f_{t+1}:=h_{t+1}'$.
    \ENDWHILE
    \end{algorithmic}
\end{algorithm}

\section{Application to Hallucination Detection}
In this Section we apply the multicalibration techniques developed in Sections \ref{sec:towards_multivalid} and \ref{sec:overfitting} to the problem of hallucination detection in LLMs. Our goal is to find a model $f(x)$ which produces (multi)calibrated confidence scores for the probability that a prompt/completion pair $x$ does not correspond to a ``hallucination''. As discussed, the key problems to solve are what to choose as the ``initial'' scoring model $f(x)$ and to determine what the groups are for data corresponding to context/completion pairs.

\subsection{The Initial Scoring Model}\label{sec:scoring}
Several methods to score hallucinations have been proposed in the literature. 
While the better the initial scoring model $f(x)$, the better we can expect our final results to be, we remark that our methodology can be applied on top of any scoring method. In this paper, we study 3 different scoring methods proposed in prior work to provide our initial score function $f(x)$, which we refine using multicalibration procedures. Note that all of the scoring functions described below are heuristics and have no calibration guarantees on their own; it is the multicalibration procedure that we use to post-process them that will endow them with guarantees.

\textbf{True/False softmax score.}
This method employs an LLM to score  the correctness of a generated answer for a specific question by asking the model whether the answer is correct and prompting it to respond with either True or False exclusively. Let $s(x)$ represent the softmax computed from the logits of the next token to be generated, and $s_k(x)$ denote the component corresponding to token $k$ \cite{kadavath2022language}. The confidence in the answer True, given the possible answers True and False, is defined as $f(x):=\tfrac{s_{\text{True}}(x)}{s_{\text{True}}(x) + s_{\text{False}}(x)}$.

\textbf{Inverse perplexity score.}
In this approach, we use an LLM to compute the output logits for a generated answer to a specific question. The function $f(x)$ is then set as the inverse perplexity of the generated answer, represented as $f(x):=\exp\left(\tfrac{1}{T-T_0}\sum_{t=T_0+1}^T\log p(x_t|x_{:t-1})\right)$, where $x$ is defined as the concatenation of question tokens $x_{1:T_0}$ and answer tokens $x_{T_0+1:T}$ \cite{jelinek1977perplexity}.

\textbf{Multiple-choice softmax score.}
This approach utilizes an LLM to assess the confidence associated with each potential answer by analyzing the output logits and outputs a score by selecting the maximum confidence amongst the available choices \cite{kadavath2022language}. To elaborate, if $s(x)$ represents the softmax for the upcoming token generation, and $s_{A_k}(x)$ signifies the component corresponding to the $k$-th answer within the set of possible answers $\{A_k\}_{k=1}^K$, the score is  defined as $f(x):=\max_{k=1,\dots,K}\tfrac{s_{A_k}(x)}{\sum_{k'=1}^K s_{A_{k'}}(x)}$.

\subsection{The Grouping Strategy}\label{sec:grouping_strategy}
Multicalibration is a guarantee parameterized by groups, and so it is important to identify ``groups'' of prompt/completion pairs $x$ that are correlated with the probability that the completion is a hallucination. When these groups effectively capture features in the data that are associated with increased likelihood of hallucination, their incorporation can lead to substantial improvements in both the algorithm's accuracy and calibration. Multicalibration techniques do not require that the groups are disjoint --- the only requirement is that, given a prompt/completion pair $x$, we are able to identify at test time which groups $x$ is a member of. This gives us the freedom to define groups from arbitrary features of the prompt, information about the user, etc. --- so long as we have the ability to determine this information in deployment. Here we discuss two strategies for defining groups. 


\textbf{Clustering.}
A natural strategy is to find semantically meaningful \emph{clusters} of prompts within some embedding space.  The clustering can potentially use information not just about the prompt, but also about the completion and the initial scoring model. To the extent that the identified clusters turn out to correlate with the likelihood of hallucination, multicalibrating with respect to groups defined by the identified clusters will improve the underlying scoring function.
Off-the-shelf text encoders and soft-clustering methods are readily applicable in this setting.

\textbf{Annotations.}
Another approach to forming groups involves using the LLM to annotate prompt/completion pairs: i.e. re-prompting the LLM with yes-or-no questions whose answers will define the groups. For example, we can ask ``Does the following question require at least high school level knowledge?'', ``Does the following prompt have to do with mathematics?'', ``Is the following prompt ambiguous?'' etc. In general, since the groups need not be disjoint, \emph{any} collection of questions can be used to induce a collection of groups. An LLM can annotate the generated text with True/False assessments, indicating whether it exhibits a set of predefined characteristics. A strength of this approach is that it provides easily human interpretable groups, and is easily extensible compared to clustering strategies. A disadvantage is that it leads to higher computational cost and latency at deployment time, and the quality of the annotations may vary with the LLM. 

\section{Experiments}\label{sec:exp}
We conduct a comprehensive experimental comparison of the methodologies introduced in Sections \ref{sec:calibration}, \ref{sec:overfitting}, \ref{sec:scoring} and \ref{sec:grouping_strategy}.

\subsection{Setup}\label{sec:exp_setup}
We conduct experiments on a range of question answering datasets, namely BigBench \cite{ghazal2013bigbench}, MMLU \cite{hendrycks2020measuring}, OpenBookQA \cite{mihaylov2018can}, TruthfulQA \cite{lin2021truthfulqa}, MathQA \cite{amini2019mathqa}, and TriviaQA \cite{joshi2017triviaqa}. These datasets enable us to assess the methods across a heterogeneous collection of queries over which the probability of hallucination varies substantially. We assess the outcomes using several state-of-the-art LLMs, namely StableBeluga-13B \cite{touvron2023llama, mukherjee2023orca}, Flan-T5-base \cite{https://doi.org/10.48550/arxiv.2210.11416}, Bloomz-7b1 \cite{muennighoff2022crosslingual}, and Mistral-7B-v0.1 \cite{jiang2023mistral}. The goal is to provide a comprehensive understanding of how these methods perform across several datasets and LLMs.

\textbf{Labeling the data.} Details in Appendix \ref{app:labeling}.

\textbf{Scoring.} Details in Appendix \ref{app:scoring_setup}.

\textbf{Grouping.} Details in Appendix \ref{app:grouping_setup}.

\subsection{Results}
\begin{table}[t!]
    \centering
    \resizebox{\columnwidth}{!}{
\begin{tabular}{lccccccc}
    \textbf{gASCE} & \textbf{uncalib.} & \textbf{IGLB} & \textbf{IGHB} & \textbf{GCULR} & \textbf{HB} & \textbf{LS} \\
    \midrule
    \textbf{Business} & 0.0645 & \textbf{0.0068} & 0.0189 & 0.0083 & 0.01 & 0.0083 \\
    \textbf{Computer Sc.} & 0.0824 & 0.0254 & 0.035 & 0.0364 & \textbf{0.0241} & 0.0366 \\
    \textbf{Engineering} & 0.1331 & \textbf{0.0523} & 0.0676 & 0.0564 & 0.0679 & 0.0562 \\
    \textbf{Ethics} & 0.1775 & \textbf{0.0189} & 0.0754 & 0.0215 & 0.0703 & 0.0214 \\
    \textbf{History} & 0.024 & 0.0195 & \textbf{0.0178} & 0.025 & 0.0239 & 0.0251 \\
    \textbf{Law} & 0.1263 & \textbf{0.0085} & 0.0422 & 0.0096 & 0.0477 & 0.0096 \\
    \textbf{Mathematics} & 0.1586 & \textbf{0.0231} & 0.0555 & 0.0254 & 0.0264 & 0.0252 \\
    \textbf{Medicine} & 0.0623 & \textbf{0.0064} & 0.0198 & 0.0069 & 0.0547 & 0.007 \\
    \textbf{Miscellaneous} & 0.0257 & 0.03 & \textbf{0.0204} & 0.0321 & 0.0349 & 0.0322 \\
    \textbf{Philosophy} & 0.0704 & \textbf{0.0181} & 0.0312 & 0.0207 & 0.028 & 0.0208 \\
    \textbf{Political Sc.} & 0.0793 & 0.0439 & 0.0425 & 0.0474 & \textbf{0.0223} & 0.0473 \\
    \textbf{Psychology} & 0.0445 & 0.0118 & 0.0144 & 0.0119 & \textbf{0.0104} & 0.0119 \\
    \textbf{Religion} & 0.0888 & 0.0643 & 0.0808 & 0.0674 & \textbf{0.033} & 0.0678 \\
    \textbf{Science} & 0.0923 & \textbf{0.0056} & 0.0244 & 0.0076 & 0.0075 & 0.0076 \\
    \textbf{Security} & 0.1492 & \textbf{0.0237} & 0.0845 & 0.0377 & 0.0329 & 0.0377 \\
    \textbf{Social Sc.} & 0.0707 & \textbf{0.0127} & 0.0296 & 0.0203 & 0.0226 & 0.0204 \\
    \bottomrule
\end{tabular}
    }
    \caption{We report the $\gASCE$ for each of the true MMLU topics, on average over different LLMs. An LLM annotation strategy is used in multicalibration methods for grouping. All methods improve the $\gASCE$ compared to before calibration. IGLB achieves best results on most groups. In particular, it performs better than GCULR on gASCE, since the first guarantees multicalibration, while the second only group-conditional unbiasedness.
    \vspace{-0.5cm}}
    \label{tab:annotation_mmlu_gasce}
\end{table}
\begin{table*}[t!]
    \centering
    \resizebox{16cm}{!}{
    \begin{tabular}{l|ccccccc}
        \textbf{MSE} & \textbf{BigBench} & \textbf{MMLU} & \textbf{OpenBookQA} & \textbf{TruthfulQA} & \textbf{MathQA} & \textbf{TriviaQA} \\
        \midrule
        \textbf{uncalib.} & 0.3242 (0.0201) & 0.3045 (0.0315) & 0.2608 (0.0037) & 0.4762 (0.135) & 0.3767 (0.0817) & 0.2802 (0.029) \\
        \textbf{IGLB} & \textbf{0.2416} (0.0027) & 0.2254 (0.0084) & 0.236 (0.0091) & \textbf{0.2016} (0.0437) & \textbf{0.1727} (0.0047) & \textbf{0.1974} (0.0308) \\
        \textbf{IGHB} & 0.2588 (0.0157) & 0.2517 (0.0138) & 0.2517 (0.0062) & 0.3051 (0.0147) & 0.1898 (0.0083) & 0.2078 (0.0299) \\
        \textbf{GCULR} & 0.2432 (0.0028) & \textbf{0.2239} (0.0083) & \textbf{0.2354} (0.009) & 0.2047 (0.0471) & 0.1728 (0.0047) & 0.1976 (0.0306) \\
        \textbf{HB} & 0.2444 (0.0018) & 0.23 (0.009) & 0.2357 (0.0094) & 0.2043 (0.041) & 0.1728 (0.0047) & 0.2026 (0.0334) \\
        \textbf{LS} & 0.2459 (0.0024) & 0.2281 (0.0093) & 0.236 (0.0091) & 0.2036 (0.0457) & \textbf{0.1727} (0.0047) & 0.2008 (0.031) \\
        \bottomrule
    \end{tabular}
    }
    \centering
    \resizebox{16cm}{!}{
    \begin{tabular}{l|ccccccc}
        \textbf{ACC.} & \textbf{BigBench} & \textbf{MMLU} & \textbf{OpenBookQA} & \textbf{TruthfulQA} & \textbf{MathQA} & \textbf{TriviaQA} \\
        \midrule
        \textbf{uncalib.} & 0.4815 (0.0443) & 0.4961 (0.0258) & 0.5506 (0.037) & 0.3333 (0.1219) & 0.3131 (0.0975) & 0.5766 (0.0372) \\
        \textbf{IGLB} & \textbf{0.5691} (0.0114) & 0.634 (0.0325) & 0.5933 (0.0356) & 0.6871 (0.1158) & \textbf{0.7779} (0.0085) & 0.7023 (0.083) \\
        \textbf{IGHB} & 0.5462 (0.0142) & 0.5858 (0.0047) & 0.5711 (0.0476) & 0.4843 (0.0655) & 0.7421 (0.0162) & 0.6781 (0.0899) \\
        \textbf{GCULR} & 0.5548 (0.0128) & \textbf{0.6381} (0.0313) & \textbf{0.5979} (0.0282) & 0.6777 (0.128) & \textbf{0.7779} (0.0085) & \textbf{0.7037} (0.0812) \\
        \textbf{HB} & 0.5613 (0.0091) & 0.6274 (0.0353) & 0.5933 (0.0447) & \textbf{0.6997} (0.1034) & \textbf{0.7779} (0.0085) & 0.6944 (0.0904) \\
        \textbf{LS} & 0.5582 (0.0169) & 0.6299 (0.034) & 0.5925 (0.0347) & 0.6698 (0.1345) & \textbf{0.7779} (0.0085) & 0.6953 (0.0896) \\
        \bottomrule
    \end{tabular}
    }
    \caption{MSE and accuracy metrics are presented for all methods across various datasets, with results displayed as the mean and standard deviation (in brackets) derived from the values produced by four LLMs. Our findings highlight the superiority of multicalibration methods, specifically IGLB and GCULR, over alternative approaches across all datasets. In particular, IGLB demonstrates a significant performance advantage over IGHB, emphasizing the effectiveness of the overfitting remedies proposed in Section \ref{sec:overfitting}.}
    \label{tab:mean_std_results}
\end{table*}

\begin{figure*}[t!]
    \centering
    \includegraphics[width=\textwidth]{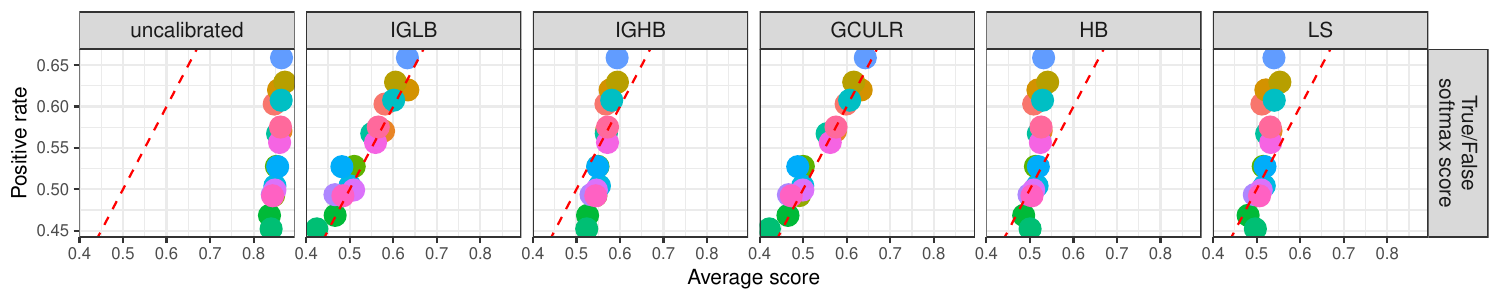}
    \vspace{-1cm}
    \caption{Average scores against accuracies across various clusters, for each method, on MMLU for StableBeluga-13B. Colors represent the groups, and the size of the points reflects their size. Multicalibration methods exhibit significantly superior alignment with the diagonal compared to standard calibration methods. In agreement with the results in Table \ref{tab:mean_std_results}, IGLB and GCULR stand out as the top performers.}
    \vspace{-0.3cm}
    \label{fig:group_rel_plot}
\end{figure*}

We conduct a comparative analysis by comparing the initial scoring functions (without any post-processing for calibration) against the same scoring functions post-processed for calibration using several algorithms: Algorithm \ref{alg:hb} (HB), Algorithm \ref{alg:ls} (LS), the logistic regression version of Algorithm \ref{alg:gcur} (GCULR), Algorithm \ref{alg:ighb} (IGHB), and Algorithm \ref{alg:iglb} (IGLB). HB and LS aim for standard (marginal) calibration and do not use any grouping strategy. GCULR produces a model satisfying group-conditional unbiasedness but not necessarily multicalibration. IGHB and IGLB produce multicalibrated models.  Compared to IGHB, IGLB implements all of the modifications discussed in Section \ref{sec:overfitting} to mitigate overfitting. All experiments in this section employ the True/False softmax score described in Section \ref{sec:scoring}. For results using other scoring methods, please refer to Appendix \ref{app:scoring_results}.

In Table \ref{tab:mean_std_results}, we present comprehensive statistics, including the mean and standard deviation (in brackets) of MSE and binary classification accuracy across the different LLMs. These metrics are analyzed across all methods and datasets. When evaluating binary classification accuracy, binary predictions are derived from the scoring function by applying a threshold of $\tfrac{1}{2}$ to the scores. This is the threshold that maximizes classification accuracy whenever the scoring function is calibrated.  The results presented here use the clustering grouping strategy. For a discussion of the results with the annotation grouping strategy, please refer to Appendix \ref{app:annotation_results}. 

Results show that IGLB and GCULR consistently out-perform across all datasets, respectively leading on MSE and accuracy. All of the models that are post-processed for calibration out-perform the initial scoring function, sometimes substantially. This underscores the importance of calibration post-processing in enhancing detection capabilities.

The results also confirm that IGHB, in its original form, is prone to overfitting. However, the modifications detailed in Section \ref{sec:overfitting} which are incorporated into IGLB, significantly enhance performance. Further insights into the specific effects of these changes are explored in Appendix \ref{app:ablation}.

Standard calibration methods such as HB and LS  perform well compared to the initial scoring function, yet consistently underperform  IGLB and  GCULR. It's noteworthy that HB achieves the best results only on OpenBookQA, possibly because of the small dataset size which causes the more complex models to overfit.

Figure \ref{fig:group_rel_plot} illustrates  average confidence scores plotted against the fraction of positive labels across various groups (each corresponding to a different color in the plot), to evaluate group-wise calibration. 
Once again we see IGLB and GCULR standing out as the top performers (represented as alignment with the diagonal). See also Appendix \ref{app:scatter_plots}.

Table \ref{tab:annotation_mmlu_gasce} provides further evidence that IGLB outperforms other methods on the gASCE evaluated using true MMLU topics. More details about this experiment in Appendix \ref{app:annotation_results}.

\section{Conclusions}
In this paper we introduce multicalibration to confidence scoring for LLMs, and develop several new techniques for both generating groups of prompts to multicalibrate with respect to, as well as new multicalibration algorithms which have improved practical performance. We show that when applied to existing scoring functions from the literature, our methods substantially improve both the error and calibration of the scores. What we have presented is an extensible framework, and so there is a clear pathway to improvement via new grouping strategies that are both semantically meaningful and correlated with LLM performance. 

\section*{Impact Statement}
 By calibrating the confidence associated with text generated by LLMs, this paper contributes to enhance trustworthiness in applications ranging from customer service interactions, to content creation and educational platforms. Calibration not only ensures more reliable and contextually appropriate responses but also mitigates the risks associated with biased or inappropriate content, thereby aligning with ethical considerations in AI development. The impact extends to critical sectors such as healthcare and finance, where the reliability of AI-generated information is of paramount importance. Furthermore, the calibration process facilitates model explainability, offering insights into decision-making mechanisms and promoting transparency in AI systems. In essence, the calibration of hallucination in LLMs is a pivotal step toward fostering responsible, trustworthy, and ethically sound AI technologies. It is important to note, however, that calibration methods should be used only with an understanding of their limitations. In our case, the provable calibration guarantees are designed to hold on prompts that are distributed as those in our calibration set are, and do not hold for adversarially generated prompts. 


\nocite{*}
\bibliography{biblio}
\bibliographystyle{icml2024}

\clearpage\appendix
\onecolumn
\section*{Appendix}


\section{On the expected calibration error}\label{app:ECE}
In our notation, the expected calibration error (ECE) is defined as 
\[ \ECE(f) := \sum_{i=1}^m\Py(f(X)\in B_i)\,\bE_\cD\Big[|\Accuracy(f) - \Confidence(f)|\Big|f(X)\in B_i\Big], \]
where $\Accuracy(f)$ is defined as $\Py(Y=\mathbbm{1}[f(X)\ge\tfrac{1}{2}])$, $B_i:=[\tfrac{i-1}{m}, \tfrac{i}{m}]$, and $\Confidence$ denotes the probability of the predicted label, that is
\[ \Confidence(f):=\max\{1-f(x), f(x)\}.\]

\section{Proof of Proposition \ref{prop:mse_decomp}}\label{app:proof_mse_decomp}
We have
\begin{align*}
    \MSE(f) &= \bE_\cD[(Y - f(X))^2]\\
    &= \bE_P[\bE_\cD[(Y - P)^2|f(X)=P]]\\
    &= \bE_P[\bE_\cD[(Y - \bE_\cD[Y|f(X)=P] + \bE_\cD[Y - P|f(X)=P])^2|f(X)=P]]\\
    &= \bE_P[\Var_\cD(Y|f(X)=P)] + \bE_P[\bE_\cD[Y - P|f(X)=P]^2]\\
    &= \bE_P[\Var_\cD(Y|f(X)=P)] + \ASCE(f).
\end{align*}

\section{Proof of Theorem \ref{thm:level set}}\label{app:proof_thm_level set}
First, we show that $\ASCE(\hf)=0$. Notice that, by construction of $\hf$, any element $p$ in the image of $\hf$ must either belong to $\grid$ or it must be such that $p=\hf(x)=f'(x)+\Delta_{p'}(f')$, for some $p'\in\grid$. In both cases, it is immediate to check that $\Delta_p(\hf)=0$. It follows that
\[ \ASCE(\hf)=\bE_{P\in\text{Im}(\hf)}[\Delta_P^2(\hf)]=0. \]
We now prove the decrease in MSE. We have
\begin{align*}
    \MSE(\hf) &= \bE_\cD[(Y - \hf(X))^2]\\
    &= \bE_\cD[(Y - f'(X) + f'(X) - \hf(X))^2]\\
    &= \MSE(f') - \sum_{p\in\grid} \Py_\cD(B_p) \Delta_{p}^2(f').
\end{align*}
Furthermore,
\begin{align*}
    \MSE(f') &= \bE_\cD[(Y - f'(X))^2]\\
    &= \bE_\cD[(Y - f(X) + f(X) - f'(X))^2]\\
    &= \MSE(f) + \sum_{p\in\grid} \Py_\cD(B_p) \bE_\cD[(f(X) - f'(X))^2 + 2(Y-f(X))(f(X)-f'(X))|B_p].
\end{align*}
Notice that $\bE_\cD[(f(X) - f'(X))^2|B_p]< \frac{1}{4m^2}=\frac{\alpha^2}{4}$. Furthermore, 
\[ \bE_\cD\left[(Y - f(X))(f(X) - f'(X))|B_p\right]\le \bE_\cD\left[|Y - f(X)|\big|B_p\right]\,\bE_\cD\left[|f(X) - f'(X)|\big|B_p\right]< 1\cdot\frac{1}{m} = \alpha. \]
It follows that $\MSE(f') < \MSE(f) + \tfrac{\alpha^2}{4} + \alpha$, which concludes the proof.

\section{Proof of Theorem \ref{thm:gcu}}\label{app:thm_gcu}
For each $g\in\cG$, we have

\begin{align*}
    \frac{\partial}{\partial \lambda_g}\MSE(\hf) &= 2\bE_\cD[\frac{\partial}{\partial \lambda_g}\hf(X)(\hf(X) - Y)]\\
    &=2\bE_\cD[g(X)(\hf(X) - Y)]\\
    &=2\bE_\cD[\hf(X) - Y|g(X)=1],
\end{align*}
where the last line follows from $g(X)\in\{0, 1\}$. Then $\tfrac{\partial}{\partial \lambda_g}\MSE(\hf)=0$ if and only if $\bE_\cD[Y-\hf(X)|g(X)=1]=0$, hence the model $\hf$ parametrized by the $\{\lambda_g\}_{g\in\cG}$ that minimizes the $\MSE$ must also satisfy group-conditional unbiasedness.

The result also holds for the cross-entropy loss
\[ \text{CrossEntropy}(\hf) := \bE_\cD[Y\log \hf(X) + (1 - Y)\log(1 - \hf(X))]. \]
Indeed,
\begin{align*}
    \left|\frac{\partial}{\partial \lambda_g}\text{CrossEntropy}(\hf)\right| &= \left|\bE_\cD\left[\frac{\partial}{\partial \lambda_g}\hf(X)\left(\frac{Y}{\hf(X)} - \frac{1-Y}{1-\hf(X)}\right)\right]\right|\\
    &=\left|\bE_\cD\left[\frac{Y - \hf(X)}{\hf(X)(1-\hf(X))}\Big|g(X)=1\right]\right|\\
    &\ge 4\left|\bE_\cD[Y-\hf(X)|g(X)=1]\right|.
\end{align*}
Hence a model $\hf$ parametrized by the $\{\lambda_g\}_{g\in\cG}$ that minimizes the cross-entropy loss must also satisfy group-conditional unbiasedness.

\section{Proof of Proposition \ref{prop:equiv_multivalid}}\label{app:equiv_multivalid}
We want to show that the following two conditions are equivalent:
\begin{align*}
    \bE_\cD[Y - f(X)|f(X)=p] &= 0\quad\forall\ p\in [0,1]\quad\quad \circled{1}\\
    \bE_\cD[Y - f(X)|f(X)\le p] &= 0\quad\forall\ p\in [0,1]\quad\quad \circled{2}
\end{align*}
First, notice that
\[ \bE_\cD[Y - f(X)|f(X)\le p] = \frac{\int_0^p \bE_\cD[Y - f(X)|f(X)=v]\,p_{f(X)}(v)\,dv}{\Py(f(X)\le p)}. \]
It follows that if $\circled{1}$ holds, then $\circled{2}$ must hold, since the expression within the integral above is 0. Vice versa, if $\circled{2}$ holds, then we must have
\[ \int_{p_1}^{p_2} \bE_\cD[Y - f(X)|f(X)=v]\,p_{f(X)}(v)\,dv = 0\quad \forall\ p_1,p_2\in[0,1]. \]
Because $\bE_\cD[Y-f(X)|f(X)=v]$ is continuous in $v$ by assumption, it follows that if there existed $p^*\in [0,1]$ such that $\bE_\cD[Y-f(X)|f(X)=p^*] \ne 0$, then there would exist a small enough interval $[p_1^*,p_2^*]\ni p^*$ where the function does not change its sign. Hence the integral above over this interval would not be zero, which would contradict $\circled{2}$.

\section{Labeling the data}\label{app:labeling}
To calibrate a scoring function, we need a calibration dataset which consists of  a collection of questions, answers, and binary labels for each answer corresponding to whether or not the answer is ``correct'' or not. While BigBench, MMLU, OpenBookQA, TruthfulQA, and MathQA conveniently contain questions, multiple answers per question, and binary labels for each answer indicating correctness, TriviaQA poses a challenge by providing only a set of correct answers for each question. To overcome this limitation, we construct labelled data for TriviaQA as follows. For each TriviaQA question, we prompt an LLM to generate four different answers. Subsequently, binary labels are assigned to these generated answers based on word overlap with the answer key provided in the dataset. After this processing, TriviaQA becomes a multiple choice question answering problem, where the multiple choice options depend on the LLM under study. The data is then randomly split into calibration and testing sets, with an 80/20 split.

\section{Scoring}\label{app:scoring_setup}
For each LLM and dataset, we experiment with each of the three initial scoring  models outlined in Section \ref{sec:scoring}. Given that our datasets consist of multiple answers per question, we select the answer with the highest score and use this score as the output of the initial model $f$.

\section{Grouping setup}\label{app:grouping_setup}
To form groups, we follow the methods described in Section \ref{sec:grouping_strategy}. For the clustering approach, text embeddings were obtained using UAE-Large-V1 \citep{li2023angle}, reduction to 20-dimensional embeddings was performed using UMAP \citep{mcinnes2018umap}, and a Gaussian mixture model was fitted on both the embeddings and one of the types of LLM scores for the most likely answer. The Bayesian Information Criterion was used to select the number of groups. We also evaluated alternative dimensionality reduction techniques such as t-SNE \citep{van2008visualizing} and PCA, as well as the use of other text encoders, obtaining conclusions similar to those described in the paper. Our methodology aligns with clustering approaches in the blindspot discovery literature \citep{eyuboglu2022domino, johnson2023does}, where clustering ensures that items in each group are semantically related and exhibit comparable classification accuracy.  

For the method that relies on annotations, our goal was to design a comprehensive taxonomy that would cover the wide range of questions appearing in the datasets. Thus, we gave StableBeluga2 questions from the different datasets and instructed it to create a classification system with groups that intersect, and then further group related categories into areas.  The resulting groups for all datasets are reported in Tables \ref{tab:taxonomy-all-datasets}. For MMLU, where the topic of each question is already provided, we asked the LLM to create a simplified classification system with fewer and non-disjoint categories based on the existing taxonomy. We also instructed it to align the original categories with the new ones and manually refined this mapping. The categories created for MMLU include those reported in Section \ref{app:annotation_results}. To obtain the annotations, we asked the model ``You are an AI designed to categorize questions accurately. The possible categories are as follows: $<$all possible categories$>$. Which of these categories does the question $<$question from dataset$>$ fall into?''. We then picked the most likely categories according to the confidence scores produced by the LLM.

\begin{table}[h]
\centering
\begin{tabular}{ll}
\hline
\textbf{Category} & \textbf{Area} \\ \hline
Humanities & Subject-Based Knowledge \\
Social Sciences & Subject-Based Knowledge \\
Natural Sciences & Subject-Based Knowledge \\
Formal Sciences & Subject-Based Knowledge \\
Professional Knowledge & Subject-Based Knowledge \\
Basic Arithmetic & Mathematical Reasoning \\
Algebra & Mathematical Reasoning \\
Geometry & Mathematical Reasoning \\
Advanced Mathematics & Mathematical Reasoning \\
Statistical and Probabilistic Reasoning & Mathematical Reasoning \\
Problem Solving & Logical and Critical Thinking \\
Inferential Reasoning & Logical and Critical Thinking \\
Analytical Reasoning & Logical and Critical Thinking \\
Common Misconceptions & Factual Accuracy and Misconceptions \\
Fact-Checking & Factual Accuracy and Misconceptions \\
Controversial and Sensitive Topics & Factual Accuracy and Misconceptions \\
Cultural Literacy & General Knowledge and Trivia \\
Historical Facts & General Knowledge and Trivia \\
Scientific Facts & General Knowledge and Trivia \\
Real-World Application & Applied Knowledge \\
Hypothetical Scenarios & Applied Knowledge \\
Cross-Disciplinary Questions & Interdisciplinary \\
Integrative Reasoning & Interdisciplinary \\ \hline
\end{tabular}
\caption{LLM-generated taxonomy of questions appearing in the datasets described in Section \ref{sec:exp_setup}.}
\label{tab:taxonomy-all-datasets}
\end{table}

\section{Results with other scoring methods}\label{app:scoring_results}
In this section, we present results similar to those in Table \ref{tab:mean_std_results}, employing the inverse perplexity and multi-choice scoring methods introduced in Section \ref{sec:scoring}. Across all datasets, the findings consistently validate the earlier observations, indicating that IGLB and GCULR perform generally better than other methods.
\begin{table*}[h!]
  \centering
  \resizebox{16cm}{!}{
  \begin{tabular}{lccccccc}
  \toprule
    & & & \textbf{Inverse perplexity score} & & & \\
    \midrule
    \midrule
    \textbf{MSE} & \textbf{BigBench} & \textbf{MMLU} & \textbf{OpenBookQA} & \textbf{TruthfulQA} & \textbf{MathQA} & \textbf{TriviaQA} \\
    \midrule
    \textbf{uncalib.} & 0.3506 (0.0104) & 0.2746 (0.0133) & 0.3506 (0.1042) & 0.22 (0.0288) & 0.2268 (0.0179) & 0.2294 (0.058) \\
    \textbf{IGLB} & 0.2441 (0.0033) & 0.2183 (0.0155) & 0.2229 (0.0417) & 0.1857 (0.0232) & \textbf{0.1991} (0.0223) & 0.1863 (0.0352) \\
    \textbf{IGHB} & 0.2593 (0.0083) & 0.2287 (0.0219) & 0.2339 (0.0447) & 0.2039 (0.0331) & 0.2062 (0.0218) & 0.1847 (0.0307) \\
    \textbf{GCULR} & \textbf{0.2439} (0.0045) & \textbf{0.2181} (0.0152) & \textbf{0.2224} (0.0403) & \textbf{0.1844} (0.0249) & 0.2001 (0.0234) & \textbf{0.1795} (0.0307) \\
    \textbf{HB} & 0.2456 (0.0024) & 0.2212 (0.0174) & 0.2213 (0.0409) & 0.1912 (0.0262) & 0.1995 (0.0226) & 0.1876 (0.0338) \\
    \textbf{LS} & 0.2473 (0.0016) & 0.223 (0.0161) & 0.2226 (0.0418) & 0.1898 (0.0207) & 0.2004 (0.0236) & 0.1917 (0.0351) \\
    \bottomrule\\
  \end{tabular}
    }
    \centering
    \resizebox{16cm}{!}{
  \begin{tabular}{lccccccc}
    \textbf{ACC.} & \textbf{BigBench} & \textbf{MMLU} & \textbf{OpenBookQA} & \textbf{TruthfulQA} & \textbf{MathQA} & \textbf{TriviaQA} \\
    \midrule
    \textbf{uncalib.} & 0.5278 (0.0315) & 0.6495 (0.0573) & 0.6156 (0.1184) & 0.6997 (0.0888) & 0.7087 (0.0448) & 0.7016 (0.0589) \\
    \textbf{IGLB} & \textbf{0.5516} (0.0152) & \textbf{0.659} (0.0499)  & \textbf{0.6232} (0.1162) & \textbf{0.7453} (0.0506) & \textbf{0.719} (0.0498)  & 0.7178 (0.0714) \\
    \textbf{IGHB} & 0.5334 (0.0234) & 0.6495 (0.0573) & 0.6156 (0.1184) & 0.706 (0.081)   & 0.7087 (0.0448) & 0.7312 (0.0606) \\
    \textbf{GCULR} & 0.5487 (0.0173) & 0.6573 (0.0509) & 0.6225 (0.115)  & \textbf{0.7453} (0.0506) & 0.7148 (0.0543) & \textbf{0.7502} (0.0707) \\
    \textbf{HB} & 0.5511 (0.0129) & 0.6497 (0.0586) & 0.6221 (0.1142) & 0.7437 (0.0494) & 0.7177 (0.0511) & 0.7132 (0.0729) \\
    \textbf{LS} & 0.5493 (0.0151) & 0.6489 (0.0576) & 0.6214 (0.1175) & \textbf{0.7453} (0.0506) & 0.7138 (0.0553) & 0.7111 (0.074)  \\
    \bottomrule\\
  \end{tabular}
    }
    \centering
    \resizebox{16cm}{!}{
  \begin{tabular}{lccccccc}
    & & & \textbf{Multiple-choice softmax score} & & & \\
    \midrule
    \midrule
    \textbf{MSE} & \textbf{BigBench} & \textbf{MMLU} & \textbf{OpenBookQA} & \textbf{TruthfulQA} & \textbf{MathQA} & \textbf{TriviaQA} \\
    \midrule
    \textbf{uncalib.} & 0.2885 (0.0144) & 0.2476 (0.0333) & 0.2203 (0.0418) & 0.2713 (0.0343) & 0.1998 (0.0189) & 0.2359 (0.0279) \\
    \textbf{IGLB} & \textbf{0.238} (0.0016)  & 0.2068 (0.0089) & \textbf{0.1985} (0.0308) & 0.2058 (0.0259) & \textbf{0.1728} (0.0042) & 0.1778 (0.0416) \\
    \textbf{IGHB} & 0.2513 (0.0028) & 0.2249 (0.0226) & 0.2175 (0.0417) & 0.2661 (0.0361) & 0.1827 (0.0124) & 0.2088 (0.0373) \\
    \textbf{GCULR} & \textbf{0.238} (0.0016)  & \textbf{0.2053} (0.0107) & 0.1986 (0.0306) & \textbf{0.2043} (0.0245) & 0.1729 (0.0041) & \textbf{0.1777} (0.0416) \\
    \textbf{HB} & 0.2407 (0.0019) & 0.2083 (0.01)   & 0.2 (0.0305)    & 0.2044 (0.0255) & 0.173 (0.0045)  & 0.1809 (0.0415) \\
    \textbf{LS} & 0.242 (0.0018)  & 0.2076 (0.0096) & 0.1984 (0.0308) & 0.2041 (0.0244) & 0.1728 (0.0044) & 0.1807 (0.0413) \\
    \bottomrule\\
  \end{tabular}
    }
    \centering
    \resizebox{16cm}{!}{
  \begin{tabular}{lccccccc}
    \textbf{ACC.} & \textbf{BigBench} & \textbf{MMLU} & \textbf{OpenBookQA} & \textbf{TruthfulQA} & \textbf{MathQA} & \textbf{TriviaQA} \\
    \midrule
    \textbf{uncalib.} & 0.5176 (0.034)  & 0.5929 (0.0587) & 0.6435 (0.0801) & 0.555 (0.0684)  & 0.7339 (0.0516) & 0.6536 (0.0602) \\
    \textbf{IGLB} & 0.5769 (0.0102) & 0.6821 (0.0148) & 0.6837 (0.0597) & 0.6824 (0.0702) & \textbf{0.7762} (0.009)  & 0.7389 (0.0891) \\
    \textbf{IGHB} & 0.5594 (0.0044) & 0.655 (0.0219)  & 0.6525 (0.0735) & 0.5566 (0.0693) & 0.7501 (0.028)  & 0.6796 (0.0598) \\
    \textbf{GCULR} & \textbf{0.5774} (0.0057) & \textbf{0.6859} (0.0155) & \textbf{0.688} (0.0564)  & 0.6824 (0.0719) & \textbf{0.7762} (0.009)  & \textbf{0.7396} (0.0884) \\
    \textbf{HB} & 0.5738 (0.0143) & 0.6791 (0.0177) & 0.685 (0.0578)  & \textbf{0.695} (0.0675)  & \textbf{0.7762} (0.009)  & 0.7363 (0.0917) \\
    \textbf{LS} & 0.5736 (0.0138) & 0.6797 (0.0165) & 0.6843 (0.0585) & 0.6855 (0.0669) & 0.776 (0.0093)  & 0.7355 (0.0924) \\
    \bottomrule
  \end{tabular}
  }
  \caption{Comparable outcomes to those presented in Table \ref{tab:mean_std_results} are reported, utilizing the inverse perplexity and multiple-choice scoring methods detailed in Section \ref{sec:scoring}. The overall findings reinforce that IGLB and GCULR consistently outperform other methods across all datasets.}
  \label{tab:otherscores}
\end{table*}

\section{Ablation study}\label{app:ablation}
In this section, we examine in isolation the impacts resulting from the modifications proposed in Section \ref{sec:overfitting}. Specifically, we conduct a comparative analysis between IGHB and IGLB against two distinct variants:

\begin{itemize}
\item $\text{\textbf{IGHB}}^{\tau}$: This variant mirrors IGHB but leverages lower and upper sets $S_{p, g}^{\tau}$ for bins, as expounded upon in Section \ref{sec:uplowsets}.
\item $\text{\textbf{IGHB}}^{\text{LS}}$: This variant employs the linear scaling patching strategy discussed in Section \ref{sec:lin_scal} instead of the constant shift patch in Algorithm \ref{alg:ighb}.
\end{itemize}
The results in Table \ref{tab:ablation} consistently reveal that IGLB, encompassing all the proposed changes to address overfitting in Section \ref{sec:overfitting}, consistently outperforms its variants, which either lack or only partially incorporate the proposed alterations. The lower and upper-level binning scheme, as introduced in Section \ref{sec:uplowsets}, emerges as the primary driver of improvement, with $\text{IGHB}^\tau$ achieving nearly comparable results to IGLB. Conversely, the implementation of linear scaling in $\text{IGHB}^{\text{LS}}$, without the alteration to the bins, results in poorer performance than IGHB alone. This observation is unsurprising, as within standard level set bins, the model remains almost constant, and linear scaling patches merely amplify the risk of overfitting.

\begin{table*}[h!]
    \centering
    \resizebox{16cm}{!}{
    \begin{tabular}{l|ccccccc}
        \textbf{MSE} & \textbf{BigBench} & \textbf{MMLU} & \textbf{OpenBookQA} & \textbf{TruthfulQA} & \textbf{MathQA} & \textbf{TriviaQA} \\
        \midrule
        \textbf{IGLB} & \textbf{0.2416} (0.0027) & \textbf{0.2254} (0.0084) & \textbf{0.236} (0.0091) & \textbf{0.2016} (0.0437) & \textbf{0.1727} (0.0047) & \textbf{0.1974} (0.0308) \\
        $\text{\textbf{IGHB}}^\tau$ & 0.2428 (0.0024) & 0.2269 (0.0096) & 0.2372 (0.0083) & 0.2043 (0.0451) & 0.173 (0.0049) & 0.1977 (0.0305)\\
        $\text{\textbf{IGHB}}^{\text{LS}}$ & 0.2521 (0.0101) & 0.2597 (0.0206) & 0.2583 (0.0051) & 0.3421 (0.0424) & 0.2 (0.0219)    & 0.2243 (0.017)\\
        \textbf{IGHB} & 0.2588 (0.0157) & 0.2517 (0.0138) & 0.2517 (0.0062) & 0.3051 (0.0147) & 0.1898 (0.0083) & 0.2078 (0.0299) \\
        \bottomrule
    \end{tabular}
    }
    \centering
    \resizebox{16cm}{!}{
    \begin{tabular}{l|ccccccc}
        \textbf{ACC.} & \textbf{BigBench} & \textbf{MMLU} & \textbf{OpenBookQA} & \textbf{TruthfulQA} & \textbf{MathQA} & \textbf{TriviaQA} \\
        \midrule
        \textbf{IGLB} & \textbf{0.5691} (0.0114) & 0.634 (0.0325) & \textbf{0.5933} (0.0356) & \textbf{0.6871} (0.1158) & \textbf{0.7779} (0.0085) & 0.7023 (0.083) \\
        $\text{\textbf{IGHB}}^\tau$ & 0.5592 (0.013) & \textbf{0.6346} (0.0347) & 0.5841 (0.0296) & 0.6698 (0.1335) & \textbf{0.7779} (0.0085) & \textbf{0.7026} (0.0827) \\
        $\text{\textbf{IGHB}}^{\text{LS}}$ & 0.5524 (0.0093) & 0.5497 (0.025) & 0.5519 (0.0404) & 0.4733 (0.0502) & 0.7398 (0.0272) & 0.6498 (0.0614)\\
        \textbf{IGHB} & 0.5462 (0.0142) & 0.5858 (0.0047) & 0.5711 (0.0476) & 0.4843 (0.0655) & 0.7421 (0.0162) & 0.6781 (0.0899) \\
        \bottomrule
    \end{tabular}
    }
    \caption{The results consistently demonstrate that IGLB outperforms its variants in the majority of cases. The principal contributor to these improvements is the lower and upper-level binning scheme implemented in $\text{IGHB}^\tau$.}
    \label{tab:ablation}
\end{table*}

\section{Scatter plots}\label{app:scatter_plots}
\begin{figure*}[h!]
    \centering
    \includegraphics[width=\textwidth]{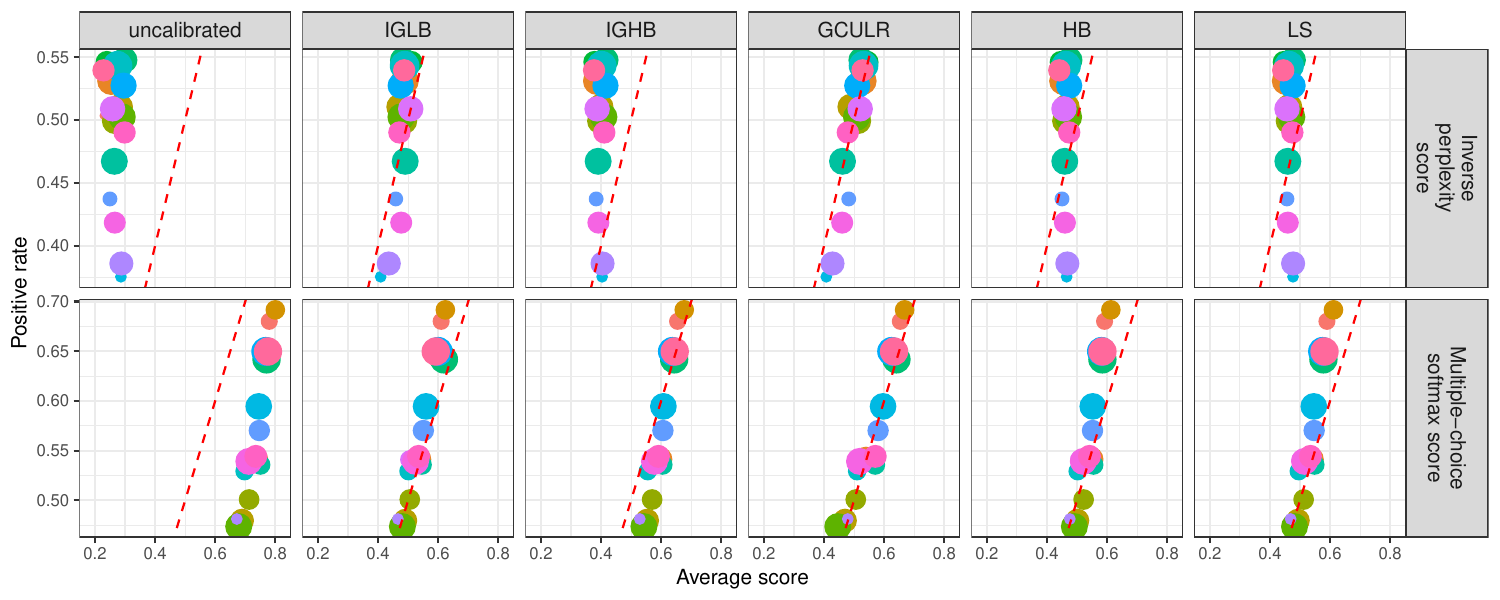}
    \caption{The average scores against the accuracy across various clusters, for each calibration method, and for inverse perplexity and multiple-choice softmax scores on MMLU and StableBeluga-13B. Conclusions are similar to those derived for Figure \ref{fig:group_rel_plot}.}
    \vspace{-0.3cm}
    \label{fig:more_scatter_plots}
\end{figure*}
Figure \ref{fig:more_scatter_plots} provides insights analogous to Figure \ref{fig:group_rel_plot}, but using inverse perplexity and multiple-choice softmax scores. All calibration methods provide post-processed scores that align significantly better with the diagonal than the intial scores before post-processing. Also in this case, multicalibration methods tend to perform better than methods such as HB and LS, which only satisfy calibration. 

\section{Results with annotations}\label{app:annotation_results}
The MMLU dataset is organized by different ``topics'' (e.g. ``Engineering'', ``Science'', etc --- see Table \ref{tab:annotation_mmlu_gasce}). The quality of a model's responses can differ substantially by topic. Here we use the LLM to attempt to annotate each prompt by topic, and then use these self-annotations as groups in our multicalibration methods. Note that the self-annotations may differ from the ``true'' groupings in the MMLU dataset because of errors in the LLM annotations or ambiguities. We can nevertheless evaluate the calibration error of each of the methods we experiment with on the \emph{true} topic groupings within the MMLU dataset. In Table \ref{tab:annotation_mmlu_gasce}, we present the mean (for standard deviation, see Appendix \ref{app:annotation_results}) of the $\gASCE$ for each true topic of MMLU. Similar to Table \ref{tab:mean_std_results}, these values are computed across the mentioned LLMs.  

We see that all methods offer improvements in calibration error compared to the uncalibrated raw scores. However, we see that the multicalibration methods which make explicit use of the self-annotated group labels substantially out-perform methods that aim for only marginal calibration. Notably, IGLB consistently achieves the lowest calibration error across nearly all groups. It is noteworthy that, while GCULR demonstrated superior accuracy in Table \ref{tab:mean_std_results}, its performance on the $\gASCE$ metric is not as impressive. This outcome aligns with theoretical expectations, as GCULR  guarantees low group-conditional bias but does \emph{not} guarantee calibration within each group, which is what we are measuring here.
\begin{table*}[t!]
    \centering
    \resizebox{16cm}{!}{
    \begin{tabular}{lccccccc}
        \textbf{gASCE} & \textbf{uncalib.} & \textbf{IGLB} & \textbf{IGHB} & \textbf{GCULR} & \textbf{HB} & \textbf{LS} \\
        \midrule
        \textbf{Business} & 0.0645 (0.0291) & \textbf{0.0068} (0.0069) & 0.0189 (0.0133) & 0.0083 (0.0056) & 0.01 (0.0048) & 0.0083 (0.0056) \\
        \textbf{Computer Sc.} & 0.0824 (0.0396) & 0.0254 (0.0139) & 0.035 (0.0113) & 0.0364 (0.0066) & \textbf{0.0241} (0.0127) & 0.0366 (0.0066) \\
        \textbf{Engineering} & 0.1331 (0.0213) & \textbf{0.0523} (0.0394) & 0.0676 (0.0323) & 0.0564 (0.0072) & 0.0679 (0.0434) & 0.0562 (0.0072) \\
        \textbf{Ethics} & 0.1775 (0.0865) & \textbf{0.0189} (0.0104) & 0.0754 (0.0716) & 0.0215 (0.009) & 0.0703 (0.0899) & 0.0214 (0.0088) \\
        \textbf{History} & 0.024 (0.0128) & 0.0195 (0.0087) & \textbf{0.0178} (0.0047) & 0.025 (0.0074) & 0.0239 (0.0121) & 0.0251 (0.0073) \\
        \textbf{Law} & 0.1263 (0.0381) & \textbf{0.0085} (0.0042) & 0.0422 (0.0254) & 0.0096 (0.003) & 0.0477 (0.0713) & 0.0096 (0.0032) \\
        \textbf{Mathematics} & 0.1586 (0.0852) & \textbf{0.0231} (0.0137) & 0.0555 (0.0119) & 0.0254 (0.0122) & 0.0264 (0.0132) & 0.0252 (0.0121) \\
        \textbf{Medicine} & 0.0623 (0.0266) & \textbf{0.0064} (0.0039) & 0.0198 (0.0122) & 0.0069 (0.0028) & 0.0547 (0.0654) & 0.007 (0.0029) \\
        \textbf{Miscellaneous} & 0.0257 (0.0091) & 0.03 (0.0269) & \textbf{0.0204} (0.0067) & 0.0321 (0.0225) & 0.0349 (0.0257) & 0.0322 (0.0225) \\
        \textbf{Philosophy} & 0.0704 (0.0285) & \textbf{0.0181} (0.0117) & 0.0312 (0.0066) & 0.0207 (0.0074) & 0.028 (0.0076) & 0.0208 (0.0071) \\
        \textbf{Political Sc.} & 0.0793 (0.0268) & 0.0439 (0.0288) & 0.0425 (0.0082) & 0.0474 (0.0229) & \textbf{0.0223} (0.0152) & 0.0473 (0.0228) \\
        \textbf{Psychology} & 0.0445 (0.0229) & 0.0118 (0.0071) & 0.0144 (0.0032) & 0.0119 (0.0051) & \textbf{0.0104} (0.0053) & 0.0119 (0.0051) \\
        \textbf{Religion} & 0.0888 (0.04) & 0.0643 (0.0337) & 0.0808 (0.0225) & 0.0674 (0.0314) & \textbf{0.033} (0.0195) & 0.0678 (0.0316) \\
        \textbf{Science} & 0.0923 (0.049) & \textbf{0.0056} (0.003) & 0.0244 (0.0098) & 0.0076 (0.0015) & 0.0075 (0.0026) & 0.0076 (0.0014) \\
        \textbf{Security} & 0.1492 (0.0526) & \textbf{0.0237} (0.0183) & 0.0845 (0.0388) & 0.0377 (0.0164) & 0.0329 (0.0329) & 0.0377 (0.0163) \\
        \textbf{Social Sc.} & 0.0707 (0.0326) & \textbf{0.0127} (0.0083) & 0.0296 (0.0214) & 0.0203 (0.0109) & 0.0226 (0.0087) & 0.0204 (0.0108) \\
        \bottomrule
    \end{tabular}
    }
    \caption{We report the $\gASCE$ obtained by each method for each of the true MMLU topics, with True/False softmax scores. An LLM annotation strategy is used in multicalibration methods for grouping. All methods improve the $\gASCE$ compared to before calibration. In particular, IGLB achieves best results almost most groups. It is meaningful to notice that, as expected from the theory, IGLB achieves better results than GCULR on gASCE, since the first guarantees multicalibration, while the second only group-conditional unbiasedness.}
    \label{tab:annotation_mmlu_gasce}
\end{table*}

\end{document}